\documentclass[letterpaper, 10 pt, journal, twoside]{IEEEtran}  

\IEEEoverridecommandlockouts                              

\usepackage{graphics} 
\usepackage{amsmath} 
\usepackage{amsmath,mathtools}
\usepackage{graphicx,subfigure} 
\usepackage{todonotes}
\usepackage{soul}

\usepackage{amssymb}
\usepackage[export]{adjustbox}
\usepackage{booktabs}
\usepackage{tabularx}
\usepackage{url}
\usepackage{tikz}

\usepackage{color}
\newcommand{\Rebuttal}[1]{\textcolor{black}{#1}}
\definecolor{orange}{RGB}{255,127,0}

\global\long\def\T{\mathbf{T}}
\global\long\def\J{\mathcal{J}}

\global\long\def\br{\mathbf{r}}

\global\long\def\bz{\mathbf{z}}

\global\long\def\be{\mathbf{e}}

\global\long\def\bR{\mathbf{R}}

\global\long\def\bl{\mathbf{l}}
\global\long\def\bx{\mathbf{x}}
\global\long\def\br{\mathbf{r}}

\global\long\def\be{\mathbf{e}}
\global\long\def\bW{\mathbf{W}}

\global\long\def\proj{\mathbf{\pi}}

\global\long\def\SO3{SO\left(3\right)}
\global\long\def\SE3{SE\left(3\right)}

\title{
Ultimate SLAM? \\ Combining Events, Images, and IMU for Robust Visual SLAM in HDR and High Speed Scenarios
}

\author{Antoni Rosinol Vidal$^*$, Henri Rebecq$^*$, Timo Horstschaefer and Davide Scaramuzza%
\thanks{Manuscript received: September, 10, 2017; Revised December, 2, 2017; Accepted December, 4, 2017.} 
\thanks{This paper was recommended for publication by Editor Francois Chaumette upon evaluation of the Associate Editor and Reviewers' comments. This work was funded by the DARPA FLA program, the Swiss National Center of Competence Research (NCCR) Robotics, through the Swiss National Science Foundation, and the SNSF-ERC starting grant.}
\thanks{$^*$Antoni Rosinol Vidal and Henri Rebecq contributed equally to this work. The authors are with the Robotics and Perception Group, Dep. of Neuroinformatics, ETH and University of Zurich, and Dep. of Informatics, University of Zurich, Switzerland.
        {\tt\small http://rpg.ifi.uzh.ch}}%
\thanks{Digital Object Identifier (DOI): see top of this page.}
}

\begin{document}

\maketitle

\begin{abstract}

Event cameras are bio-inspired vision sensors that output pixel-level brightness changes instead of standard intensity frames.
These cameras do not suffer from motion blur and have a very high dynamic range, which enables them to provide reliable visual information during high speed motions or in scenes characterized by high dynamic range.
However, event cameras output only little information when the amount of motion is limited, such as in the case of almost still motion.
Conversely, standard cameras provide instant and rich information about the environment most of the time (in low-speed and good lighting scenarios), but they  fail severely in case of fast motions, or difficult lighting such as high dynamic range or low light scenes.
In this paper, we present the first state estimation pipeline that leverages the complementary advantages of these two sensors by fusing in a tightly-coupled manner events, standard frames, and inertial measurements.
We show on the publicly available Event Camera Dataset that our \emph{hybrid} pipeline leads to an accuracy improvement of 130\% over \emph{event-only} pipelines, and 85\% over \emph{standard-frames-only} visual-inertial systems, while still being computationally tractable.
Furthermore, we use our pipeline to demonstrate---to the best of our knowledge---the first autonomous quadrotor flight using an event camera for state estimation, unlocking flight scenarios that were not reachable with traditional visual-inertial odometry, such as low-light environments and high-dynamic range scenes.

\end{abstract}


\section*{Supplementary Material}
Videos of the experiments: \url{http://rpg.ifi.uzh.ch/ultimateslam.html}

\section{Introduction}
\label{sec:introduction}

\IEEEPARstart{T}{e} task of estimating a sensor's ego-motion has important applications in various fields, such as augmented/virtual reality or autonomous robot control.
In recent years, great progress has been achieved using visual and inertial information (\cite{Mourikis07icra,Leutenegger15ijrr,Forster17troOnmanifold}).
However, due to some well-known limitations of traditional cameras (motion blur and low dynamic-range), these Visual Inertial Odometry (VIO) pipelines still struggle to cope with a number of situations, such as high-speed motions or high-dynamic range scenarios.

\begin{figure}[htp]
\includegraphics[width=1.0\columnwidth]{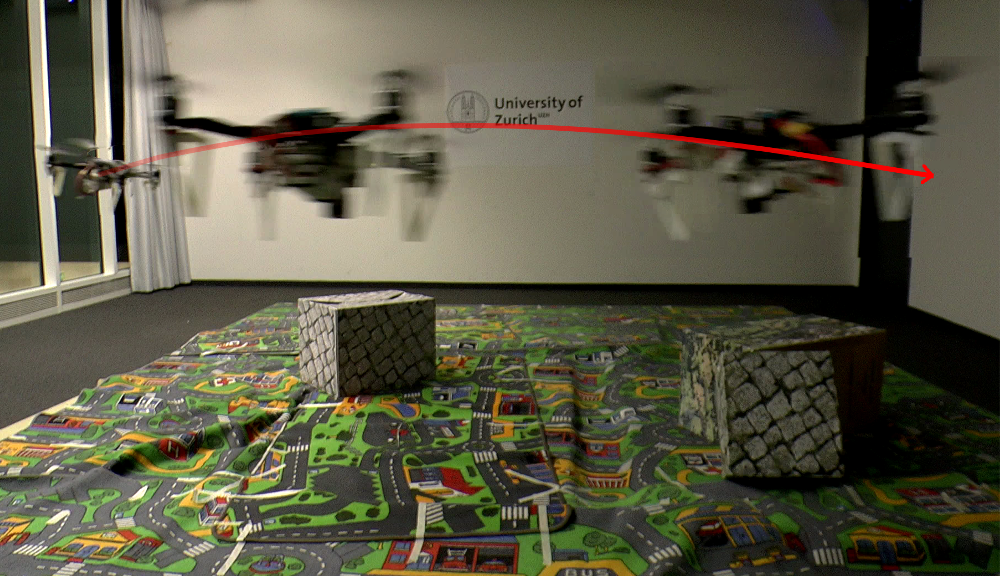}
\begin{minipage}[b][][s]{\columnwidth}
  \vspace{0.4em}
  \centering
  \resizebox{\columnwidth}{!}{%
    \setlength\tabcolsep{1.5pt} 
    \begin{tabular}{ccc}
    Standard Frame &
    Event Frame &
    Events \\
    \includegraphics[width=.325\columnwidth]{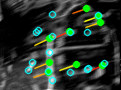} &
    \includegraphics[width=.325\columnwidth]{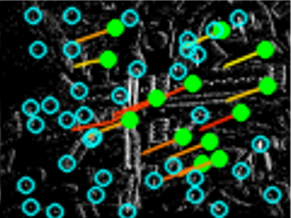} &
    \includegraphics[width=.325\columnwidth]{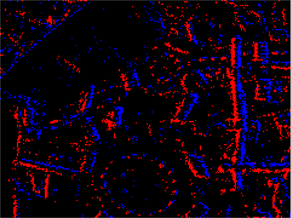}
  \end{tabular}%
  }
\end{minipage}

\caption{Our state estimation pipeline combines events, standard frames, and inertial measurements to provide robust state estimation, and can run onboard an autonomous quadrotor with limited computational power. Bottom Left: Standard frame, Bottom Middle: Virtual event frame, Bottom Right: Events only (blue: positive events, red: negative events).}
\label{fig:eye_catcher}
\end{figure}

Novel types of sensors, called event cameras, offer great potential to overcome these issues.
Unlike standard cameras, which transmit intensity frames at a fixed framerate, event cameras, such as the Dynamic Vision Sensor (DVS) ~\cite{Lichtsteiner08ssc}, only transmit \emph{changes of intensity}.
Specifically, they transmit per-pixel intensity changes at the time they occur, in the form of a set of asynchronous \emph{events}, where each event carries the space-time coordinates of the brightness change, and its sign.

Event cameras have numerous advantages over standard cameras: a latency in the order of microseconds and a very high dynamic range (140 dB compared to 60 dB of standard cameras).
Most importantly, since all the pixels capture light independently, such sensors do not suffer from motion blur.

Event cameras transmit, in principle, all the information needed to reconstruct a full video stream \cite{Cook11ijcnn,Bardow16cvpr,Reinbacher16bmvc}, and one could argue that an event camera alone is sufficient to perform state estimation.
In fact, this has been shown recently in \cite{Rebecq17ral} and \cite{Kim16eccv}.
However, to overcome the lack of intensity information, these approaches need to reconstruct, in parallel, a consistent representation of the environment (a semi-dense depth map in \cite{Rebecq17ral} or a dense depth map with intensity values in \cite{Kim16eccv}), by combining---in one way or another---information from a large number of events to recover most gradients in the scene.

Conveniently, standard cameras provide direct access to intensity values, but do not work in low-light conditions, suffer from motion blur during fast motions (due to the synchronous exposure on the whole sensor), and have a limited dynamic range (60 dB), resulting in frequent over- or under-exposed areas in the frame.

Observing this complementarity, in this paper we propose a pipeline that leverages the advantages of both sensing modalities in combination with an inertial measurement unit (IMU) to yield a robust, yet accurate, state estimation pipeline.

While there is a considerable body of literature investigating the use of standard cameras with an IMU to perform state estimation, as well as recent work using an event camera with an IMU, combining all three sensing modalities is yet an open problem.
Additionally, in the core application that we envision---flying autonomously a quadrotor with an event camera---there is no specific literature, although attempts to use an event camera for quadrotor flight can be traced to a single paper \cite{Hordijk17arxiv}, which is currently limited to vertical landing maneuvers.

In this work, we propose---to the best of our knowledge---the first state estimation pipeline that fuses all three sensors, and we build on top of it to propose the first quadrotor system that can advantageously exploit this hybrid sensor combination to fly in difficult scenarios, using only onboard sensing and computing (Fig.~\ref{fig:eye_catcher}).

\subsection*{Contributions}

A frontal comparison with state-of-the-art, commercial visual-inertial pipelines (like for example the ones used for the Snapdragon flight \cite{Qualcomm16SnapdragonFlight} or Google Tango \cite{Tango}) is not our goal in this work.
Indeed, such solutions typically use one or more high quality cameras with a much higher resolution than the sensor we used, and are carefully engineered to work well in the most common consumer situations.
Instead, in this work, we focus on difficult scenarios, and show, for the first time, that (i) it is possible to run state estimation with an event camera onboard a computationally limited platform, and (ii) we show that it can unlock, in a set of difficult scenarios, the possibility for autonomous flight where even commercial systems would struggle.

Specifically, our contributions in this paper are three-fold:
\begin{itemize}
  \item We introduce the first state estimation pipeline that fuses events, standard frames, and inertial measurements to provide robust and accurate state estimation.
While our pipeline is based on \cite{Rebecq17bmvc}, we extend it to include standard frames as an additional sensing modality, and propose several improvements to make it usable for real-time applications, with a focus on mobile robots.
  \item We evaluate quantitatively the proposed approach and show that using standard frames as an additional modality improves the accuracy of state estimation while keeping the computational load tractable.
  \item We show that our method can be applied for state estimation onboard an autonomous quadrotor, and demonstrate in a set of experiments that the proposed system is able to fly reliably in challenging situations, such as low-light scenes or fast motions.
\end{itemize}

Our work aims at highlighting the potential that event cameras have for robust state estimation, and we hope that our results will inspire other researchers and industries to push this work forward, towards the wide adoption of event cameras on mobile robots.

The rest of the paper is organized as follows: section \ref{sec:related_work} reviews related literature on event-based ego-motion estimation methods, particularly those involving event cameras.
In section \ref{sec:state_estimation}, we present our hybrid state estimation pipeline that fuses events, standard frames and inertial measurements in a tightly-coupled fashion, and evaluate it quantitatively on the publicly available Event Camera Dataset \cite{Mueggler17ijrr}.
Section \ref{sec:quadrotor_flight} describes how the proposed approached can be used to fly a quadrotor autonomously, and demonstrate in a set of real-life experiments that it unlocks challenging scenarios difficult to address with traditional sensing \ref{sec:flight_experiments}.
Finally, we draw conclusions in section~\ref{sec:conclusions}.

\section{Related Work}
\label{sec:related_work}

Using visual and inertial sensors for state estimation has been extensively studied over the past decades.
While the vast majority of these works use standard cameras together with an IMU, a recent parallel thread of research that uses event cameras in place of standard cameras has recently flourished.

\vspace{0.65ex}
\paragraph{Visual-inertial Odometry with Standard Cameras}

The related work on visual-inertial odometry (VIO) can be roughly segmented into three different classes, depending on the number of camera poses that are used for the estimation.
While full smoothers (or batch nonlinear least-squares algorithms) estimate the complete history of poses, fixed-lag smoothers (or sliding window estimators) consider a window of the latest poses, and filtering approaches only estimate the latest state.
Both fixed-lag smoothers and filters marginalize older states and absorb the corresponding information in a Gaussian prior. More specifically:
\begin{itemize}
  \item Filtering algorithms enable efficient estimation by restricting the inference process to the latest state of the system.
  A example approach of a filter-based visual-inertial odometry system is \cite{Jones11ijrr}.
  \item Fixed-lag smoothers estimate the states that fall within a given time window, while marginalizing out older states, as for example, \cite{Leutenegger15ijrr}.
  \item Full smoothing methods estimate the entire history of the states (camera trajectory and 3D landmarks), by solving a large nonlinear optimization problem.
  A recent approach in this category was proposed by \cite{Forster17troOnmanifold}.
\end{itemize}

\vspace{0.65ex}
\paragraph{Visual-inertial Odometry with Event Cameras}
Since the introduction of the first commercial event camera in 2008 \cite{Lichtsteiner08ssc}, event cameras have been considered for state estimation by many different authors.
While early works focused on addressing restricted and easier instances of the problem, like rotational motion estimation (\cite{Cook11ijcnn}, \cite{Kim14bmvc}, \cite{Gallego17ral}, \cite{Reinbacher17iccp}), or Simultaneous Localization and Mapping (SLAM) in planar scenes only \cite{Weikersdorfer13icvs}, it has been shown recently that 6-DOF pose estimation using only an event camera is possible (\cite{Rebecq17ral,Kim16eccv}). 

In parallel, other authors have explored the use of complementary sensing modalities, such as a depth sensor \cite{Weikersdorfer2014icra}, or a standard camera (\cite{Censi14icra}, \cite{Kueng16iros}).
However, (i) none of these image-based pipelines make use of inertial measurements, and (ii) both of them use the intensity of the frames as a template, to which they align the events.
Therefore, these approaches work only when the standard frames are of good quality (sharp and correctly exposed); they will fail in those particular cases where the event camera has an advantage over a standard camera (high-speed motions, and HDR scenes).

Using an event camera and an IMU has only been explored very recently.
\cite{Mueggler17arxiv} showed how to fuse events and inertial measurements into a continuous time framework, but their approach is not suited for real-time usage because of the expensive optimization required to update the spline parameters upon receiving every event.
\cite{Zhu17cvpr} proposed to track a set of features in the event stream using an iterative Expectation-Maximization scheme that jointly refines each feature's appearance and optical flow, and then fuse these tracks using an Extended Kalman Filter to yield an event-based visual-inertial odometry pipeline.
Unfortunately, due to the expensive nature of their feature tracker, the authors of \cite{Zhu17cvpr} reported that their pipeline cannot run in real-time in most scenarios.

In \cite{Rebecq17bmvc}, we proposed an accurate event-based visual inertial odometry pipeline that can run in real-time, even on computationally limited platforms, such as smartphone processors.
The key of this approach was to estimate the optical flow generated by the camera's rigid body motion by exploiting the current camera pose, scene structure, and inertial measurements.
We then efficiently generated virtual, motion-compensated event frames using the computed flow, and further tracked visual features across multiple frames.
Those feature tracks were finally fused with inertial information using keyframe-based nonlinear optimization, in the style of \cite{Leutenegger15ijrr} and \cite{Forster17troOnmanifold}.
While our proposed state estimation approach is strongly inspired by this work (i.e., \cite{Rebecq17bmvc}), we extend it by allowing it to additionally work with frames from a standard camera, and propose several changes to the pipeline to adapt it to run onboard a flying robot.

\vspace{0.65ex}
\paragraph{Quadrotor Control with an Event Camera}

Although the research on robot control with event cameras is still in its infancy, previous work has demonstrated possible interesting applications.
\cite{Mueggler14iros} mounted a DVS sensor on a quadrotor and showed that it can be used to track the 6-DOF motion of a quadrotor performing a high speed flip maneuver, although the tracker only worked for an artificial scene containing a known black square painted over a white wall. Also, the state estimation was performed offline, and therefore not used for closed-loop control of the quadrotor.
More recently, \cite{Hordijk17arxiv} showed closed-loop take-off and landing of a quadrotor using an event camera.
Their system, however, relied on computing optical flow and assumed the flow field to be divergent, thus it cannot be used for general 6-DOF control of a quadrotor, unlike our approach.

\section{Hybrid State Estimation Pipeline}
\label{sec:state_estimation}

Our proposed state estimation pipeline is largely based on \cite{Rebecq17bmvc}.
However, while \cite{Rebecq17bmvc} used only an event camera combined with an IMU, we propose to allow for an additional sensing modality: a standard camera, providing intensity frames at a fixed framerate.
For this reason, we focus below on describing the differences between our approach and \cite{Rebecq17bmvc} in order to also consider standard frames.
Finally, we evaluate the improved pipeline on the Event Camera Dataset \cite{Mueggler17ijrr} and show evidence that incorporating standard frames in the pipeline leads to an accuracy boost of 130\% over a pipeline that uses only events plus IMU, and 85\% over a pipeline that uses only standard frames plus IMU.

\subsection{Overview}

\cite{Rebecq17bmvc} can be briefly summarized as follows.
The main idea is to synthesize virtual frames (\emph{event frames}) from spatio-temporal windows of events, and then perform feature detection and tracking using classical computer vision methods, namely the FAST corner detector \cite{Rosten06eccv} and the Lucas-Kanade tracker \cite{Lucas81ijcai}.
Feature tracks are used to triangulate the 3D locations of the corresponding landmarks whenever it can be done reliably.
Finally, the camera trajectory and the positions of the 3D landmarks are periodically refined by minimizing a cost function involving visual terms (reprojection error) and inertial terms, thus effectively fusing visual and inertial information.

In this paper, we propose to not only maintain feature tracks from virtual event frames, but to also maintain, in parallel, feature tracks from standard frames as well.
We then feed the feature tracks coming from these two heterogeneous sources (virtual event frames and standard frames) to the optimization module, thus effectively refining the camera poses using the events, the standard frames, and the IMU.

\vspace{0.45ex}
\subsubsection{Coordinate Frame Notation}

A point $P$ represented in a coordinate frame $A$ is written as position vector $\prescript{}{A}{\br_P}$.
A transformation between frames is represented by a homogeneous matrix $\T_{AB}$ that transforms points from frame $B$ to frame $A$. Its rotational part is expressed as a rotation matrix $\bR_{AB} \in \SO3$. 
Our algorithm uses a hybrid sensor composed of an event camera, a standard camera, and an IMU rigidly mounted together.
The sensor body is represented relative to an inertial world frame $W$.
\Rebuttal{Within the sensor body}, we distinguish the event camera frame $C_0$, the standard camera frame $C_1$ and the IMU-sensor frame $S$.
To obtain $\T_{SC_0}$ and $\T_{SC_1}$, an extrinsic calibration of the \{event camera + standard camera + IMU system\} must be performed.

\vspace{0.45ex}
\subsubsection{Spatio-temporal Windows of Events}

\begin{figure}
\centering
\includegraphics[width=1.0\columnwidth]{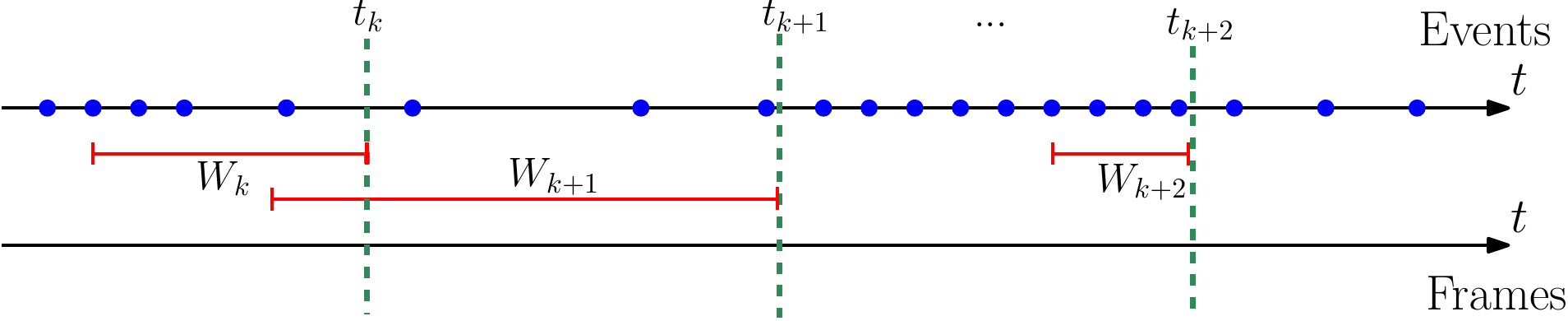}
\caption{Upon receiving a new frame from the standard camera at time $t_k$, we select a spatio-temporal window of events $W_k$, containing a fixed number of events ($N = 4$ in this example). Note that the temporal size of each window is automatically adapted to the event rate. Blue dots correspond to events, and the dashed green lines correspond to the times at which standard frames are received. The bounds of the spatio-temporal windows of events considered are marked in red.}

\label{fig:spatial_temporal_windows}
\end{figure}

We synchronize the spatio-temporal windows of events on the timestamps of the standard frames.
Upon reception of each standard frame at time $t_k$, a new spatio-temporal window of events $W_k$ is created (Fig.~\ref{fig:spatial_temporal_windows}).
The $k^{th}$ window is defined as the set of events $W_k = \left\lbrace e_{j(t_k)-N+1}, ..., e_{j(t_k)} \right\rbrace$, where $j(t_k)$ is the index of the first event whose timestamp $t_j < t_k$, and $N$ is the window size parameter.
Note that the duration of each window is inversely proportional to the event rate.

\vspace{0.45ex}
\subsubsection{Synthesis of Motion-Compensated Event Frames}
As in \cite{Rebecq17bmvc}, we then collapse every spatio-temporal window of events to a synthetic event frame $I_k$ by drawing each event on the image plane, after correcting for the motion of each event according to its individual timestamp.

Let 
$I_k(\bx) = \sum_{e_i \in W_k}{\delta(\bx - \bx'_i)},$
where \Rebuttal{function ${\delta(\bx)}$ is the Kronecker delta,} $\bx'_i$ is the \emph{corrected} event position, obtained by transferring event $e_i$ to the reference \Rebuttal{event} camera frame \Rebuttal{${C_0}_k$}:
\begin{equation}
\label{eq:event_back_projection}
\bx'_i = \proj_0 ( T_{t_k,t_i} ( Z(\bx_i) \proj_0^{-1}(\bx_i) )),
\vspace{-0.5ex}
\end{equation}
where $\bx_i$ is the pixel location of event $e_i$, $\proj_0 \left( . \right)$ the event camera projection model, obtained from prior intrinsic calibration, and $T_{t_l,t_m}$ the incremental transformation between the camera poses at times $t_l$ and $t_m$, obtained through integration of the inertial measurements (we refer the reader to \cite{Rebecq17bmvc} for details).
\Rebuttal{$Z(\bx_i)$ is the scene depth at time $t_i$ and pixel $\bx_i$, which we can estimate using 2D linear interpolation (on the image plane) of the landmarks reprojected on the current camera frame ${C_0}_i$.
In practice, as in \cite{Rebecq17bmvc}, we observed that using the median depth of the landmarks in the field of view instead of linearly interpolating the depth gives satisfactory results at a lower computational cost.}
The quality of the motion compensation depends on the quality of the 3D landmarks available, therefore the quality of the event frames improves when also using the landmarks from the standard frames.

The number of events $N$ in each spatio-temporal window is a parameter that needs to be adjusted depending on the amount of texture in the scene.
As an example, for the quadrotor experiments presented in section \ref{sec:quadrotor_flight}, we used $N = 20~000$ events per frame.

\vspace{0.45ex}
\subsubsection{Feature Tracking}

We use the FAST corner detector to extract features \cite{Rosten06eccv}, both on the virtual event frames, and the standard camera frames.
Those features are then tracked independently across standard frames and event frames using the KLT tracker \cite{Lucas81ijcai} (see Fig.~\ref{fig:eye_catcher}).
This yields two sets of independent features tracks $\left\lbrace \bz^{0,j,k} \right\rbrace$, $\left\lbrace \bz^{1,j,k} \right\rbrace$ (where $j$ is the feature track index, and $k$ is the frame index).
For each sensor, each feature is treated as a \emph{candidate} feature, and tracked over multiple frames.
Once a feature can be triangulated reliably, the corresponding 3D landmark is triangulated through linear triangulation \cite{Hartley03book}, and converted to a \emph{persistent} feature which will be further tracked across the next frames.
We re-detect features on each sensor as soon as the number of tracked features falls below a threshold.
We used the same detection and tracking parameters for the motion-compensated event frames and for the standard frames.
The FAST threshold we used was $50$.
We used a a pyramidal implementation of KLT with $2$ pyramid levels, and a patch size of $24 \times 24$ pixels.
Additionally, we used a bucketing grid (where each grid cell has size $32 \times 32$ pixels) to ensure that features are evenly distributed in each sensor's image plane.


\vspace{0.45ex}
\subsubsection{Visual-inertial Fusion through Nonlinear Optimization}

The visual-inertial localization and mapping problem is formulated as the joint optimization of a cost function that contains three terms: two weighted reprojection errors corresponding respectively to the observations from the event camera and the standard camera, plus an inertial error term~$\be_s$:
\vspace{-0.5em}
$$J = \sum_{i=0}^{1}\sum_{k=1}^{K}\sum_{j \in \J(i,k)}{{\be^{i,j,k}}^T \bW_r^{i,j,k} \be^{i,j,k}} + \sum_{k=1}^{K-1}{{\be_s^k}^T \bW_s^k \be_s^k}$$
where $i$ denotes the sensor index, $k$ denotes the frame index, and $j$ denotes the landmark index. The set $\J(i,k)$ contains the indices of landmarks maintained in the $k^{th}$ frame by sensor $i$. Additionally, $W_r^{i,j,k}$ is the information matrix of the landmark measurement $\bl_{i,j}$, and $W_s^k$ that of the $k^{th}$ IMU error.
The reprojection error is:
$$\be_r^{i,j,k} = \bz^{i,j,k} - \proj_i \left( \T_{C_iS}^k \T_{SW}^k \bl^{i,j} \right)$$ where $\bz^{i,j,k}$ is the measured image coordinate of the $j^{th}$ landmark on the $i^{th}$ sensor at the $k^{th}$ frame.
We use standard IMU kinematics and biases model (see \cite{Forster17troOnmanifold} for example) to predict the current state based on the previous state.
Then, the IMU error terms are computed as the difference between the prediction based on the previous state and the actual state.
For orientation, a simple multiplicative minimal error is used.
For details, we refer the reader to \cite{Leutenegger15ijrr}.

The optimization is carried out not on all the frames observed but on a bounded set of frames composed of $M$ keyframes (we use the same keyframe selection criterion as \cite{Rebecq17bmvc}), and a sliding window containing the last $K$ frames.
In between frames, the prediction for the sensor state is propagated using the IMU measurements.
We employ the Google Ceres \cite{ceres-solver} optimizer to carry out the optimization.

\Rebuttal{Notice that with this formulation we avoid an explicit switching policy between standard and event camera: the optimization naturally uses the best sensing modalities available.}

\subsubsection{Additional Implementation Details}

\paragraph{Initialization}

We assume that the sensor remains static during the initialization phase of the pipeline, during one or two seconds.
We collect a set of inertial measurements and use them to estimate the initial attitude (pitch and roll) of the sensor, as well as to initialize the gyroscope and accelerometer biases.
\paragraph{No-Motion Prior for Almost-Still Motions}

When the sensor is still, no events are generated (except noise events).
To handle this case in our pipeline, we add a strong zero velocity prior to the optimization problem whenever the event rate falls below a threshold, thus forcing the sensor to be still.
We used a threshold in the order of $10^{3}$ events/s in our experiments, and measured the event rate using windows of $20$ ms.

\subsection{Evaluation}
\label{sec:quantitative_evaluation}

\begin{table}[]
\centering
\begin{minipage}{0.5\textwidth}
\resizebox{\textwidth}{!}{
\begin{tabular}{l p{1.0cm} p{1.0cm} p{1.0cm} p{1.0cm} p{1.0cm} p{0.8cm}}
\hline
Sequence             & \multicolumn{2}{c}{\textbf{Proposed (\Rebuttal{Fr + E + I})}} & \multicolumn{2}{c}{\Rebuttal{\textbf{E + I}}}  & \multicolumn{2}{c}{\Rebuttal{\textbf{Fr + I}}}          \\ \hline
                     & Mean Position Error (\%)  & Mean Yaw Error (deg/m) & Mean Position Error (\%) & Mean Yaw Error (deg/m) & Mean Position Error (\%)  & Mean Yaw Error (deg/m) \\
              boxes\_6dof & \textbf{{0.30}} & \textbf{{0.04}} & 0.44 & 0.05 & 0.30 & 0.06 \\
       boxes\_translation & 0.27 & \textbf{{0.02}} & 0.76 & 0.05 & \textbf{{0.17}} & 0.03 \\
            dynamic\_6dof & \textbf{{0.19}} & 0.10 & 0.38 & \textbf{{0.06}} & 0.62 & 0.10 \\
     dynamic\_translation & \textbf{{0.18}} & \textbf{{0.15}} & 0.59 & 0.16 & 0.67 & 0.26 \\
               hdr\_boxes & \textbf{{0.37}} & \textbf{{0.03}} & 0.67 & 0.09 & 0.78 & 0.17 \\
              hdr\_poster & 0.31 & 0.05 & 0.49 & \textbf{{0.04}} & \textbf{{0.28}} & 0.08 \\
             poster\_6dof & \textbf{{0.28}} & \textbf{{0.07}} & 0.30 & 0.08 & 0.59 & 0.11 \\
      poster\_translation & \textbf{{0.12}} & 0.04 & 0.15 & \textbf{{0.04}} & 0.23 & 0.08 \\
             shapes\_6dof & \textbf{{0.10}} & \textbf{{0.04}} & 0.48 & 0.06 & 0.17 & 0.05 \\
      shapes\_translation & \textbf{{0.26}} & 0.06 & 0.41 & \textbf{{0.04}} & 0.29 & 0.11 \\

\hline
\end{tabular}%
}
\caption{Accuracy of the proposed approach using frames (Fr),  events (E) and IMU (I), against using events and IMU, and using frames and IMU.}
\label{tab:accuracy_comparison}
\end{minipage}
\end{table}

\begin{table}[]
\centering
\begin{minipage}{0.5\textwidth}
\resizebox{\textwidth}{!}{
\begin{tabular}{l p{1.4cm} p{1cm} p{1.4cm} p{1cm}}
\hline
Sequence             & \multicolumn{2}{c}{\textbf{Proposed (\Rebuttal{Fr + E + I})}}  & \multicolumn{2}{c}{\textbf{State-of-the-art (\Rebuttal{E + I}) \cite{Rebecq17bmvc}}}           \\ \hline
                     & Mean Position Error (\%) & Mean Yaw Error (deg/m) & Mean Position Error (\%)  & Mean Yaw Error (deg/m) \\
             boxes\_6dof & \textbf{{0.30}} & \textbf{{0.04}} & 0.36 & 0.11 \\
       boxes\_translation & \textbf{{0.27}} & \textbf{{0.02}} & 0.31 & 0.08 \\
            dynamic\_6dof & \textbf{{0.19}} & \textbf{{0.10}} & 0.56 & 0.41 \\
     dynamic\_translation & \textbf{{0.18}} & 0.15 & 0.39 & \textbf{{0.06}} \\
               hdr\_boxes & \textbf{{0.37}} & \textbf{{0.03}} & 0.59 & 0.20 \\
              hdr\_poster & \textbf{{0.31}} & \textbf{{0.05}} & 0.33 & 0.19 \\
             poster\_6dof & \textbf{{0.28}} & \textbf{{0.07}} & 0.40 & 0.16 \\
      poster\_translation & \textbf{{0.12}} & \textbf{{0.04}} & 0.46 & 0.10 \\
             shapes\_6dof & \textbf{{0.10}} & \textbf{{0.04}} & 0.42 & 0.18 \\
      shapes\_translation & \textbf{{0.26}} & \textbf{{0.06}} & 0.50 & 0.13 \\
\hline
\end{tabular}%
}
\caption{Accuracy of the proposed approach using frames (Fr), events (E) and IMU (I), against \cite{Rebecq17bmvc}, which uses events and IMU.}
\label{tab:accuracy_comparison_zevio}
\end{minipage}
\end{table}

We evaluate the proposed pipeline quantitatively on the Event Camera Dataset \cite{Mueggler17ijrr}, which features various scenes with ground truth tracking information. In particular, it contains extremely fast motions and scenes with very high dynamic range, recorded with the DAVIS\footnote{\url{https://inilabs.com/products}} \cite{Brandli14ssc} sensor. \Rebuttal{As in \cite{Zhu17cvpr}, we only use the datasets from the Event Camera Dataset that are relevant for Visual-Inertial Odometry. Specifically, we exclude the rotational only datasets, as well as the datasets without inertial measurements.}

The DAVIS sensor embeds a $240\times180$ pixels event camera with a 1kHz IMU and also delivers standard frames at 24Hz.
Events, standard frames, and IMU measurements are synchronized on hardware.
The IMU is delayed by a constant time offset in the order of $2.5$ms compared to the events and standard frames (because of the low-pass filter of the IMU). We estimated this delay using Kalibr \cite{Furgale13iros}.

To evaluate the results, the estimated and ground truth trajectories are aligned with a 6-DOF transformation in SE3, using 5 seconds of the trajectory (starting at second 3 and ending at second 8).
Then, we compute the mean position error (Euclidean distance) and the yaw error as percentages of the total traveled distance. Due to the observability of the gravity direction, the error in pitch and roll is constant and comparable for each pipeline. Thus we omit them for compactness.

\begin{figure*}[t]
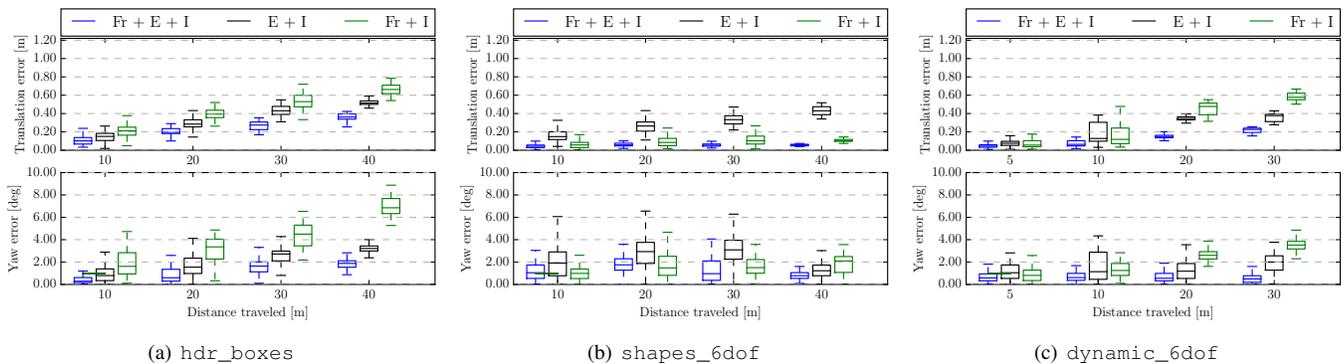

  \centering     

  \subfigure[\texttt{hdr\_boxes}]{\label{fig:event_frame_few_events}\includegraphics[trim={0 0 0 0cm},clip,width=0.325\textwidth]{results/hdr_boxes}}
  \subfigure[\texttt{shapes\_6dof}]{\includegraphics[trim={0 0 0 0cm},clip,width=0.325\textwidth]{results/shapes_6dof}}
  \subfigure[\texttt{dynamic\_6dof}]{\includegraphics[trim={0 0 0 0cm},clip,width=0.325\textwidth]{results/dynamic_6dof}}

  \caption{Comparison of the proposed approach, using frames (Fr), events (E), and IMU (I), on three datasets from the Event Camera Dataset~\cite{Mueggler17ijrr}. The graphs show the relative errors measured over different segments of the trajectory as proposed in \cite{Geiger12cvpr}. Additional plots for all the datasets are provided in the supplementary material.
  }
  \label{fig:boxplots}
\end{figure*}

Table \ref{tab:accuracy_comparison} shows the results obtained when running the pipeline in its proposed mode, using \Rebuttal{standard frames (Fr), events (E), and IMU (I)}. To further quantify the accuracy gained by using events and frames (plus IMU), compared to using only events or only frames (plus IMU), we run our proposed pipeline using the three different combinations, and report the results in Table \ref{tab:accuracy_comparison}.
Additionally, in Fig.~\ref{fig:boxplots}, we use the relative error metrics proposed in \cite{Geiger12cvpr}, which evaluate the relative error by averaging the drift over trajectories of different lengths.
Using jointly standard frames, events and IMU leads to an average position accuracy improvement of $85\%$ compared to using frames and IMU only, and $130\%$ against using events and IMU only.
\Rebuttal{Notice that the Event Camera Dataset was made to showcase the situations where an event camera would be more useful. Nevertheless, datasets like boxes\_translation and shapes\_6dof show that using standard frames might still be advantageous compared to using only events, as can be seen in Table \ref{tab:accuracy_comparison} and the detailed analysis in the supplementary material.}

Table \ref{tab:accuracy_comparison_zevio} provides a comparison between our approach and the state-of-the-art \cite{Rebecq17bmvc}.

\Rebuttal{The events plus IMU pipeline in Table \ref{tab:accuracy_comparison} is not the same as \cite{Rebecq17bmvc} in Table \ref{tab:accuracy_comparison_zevio}; the former generates event frames at a fixed rate, while the latter generates them at a rate that depends on the event rate, we refer the reader to \cite{Rebecq17bmvc} for further details. Moreover, the parameters both pipelines share do not necessarily have the same values. These reasons account for the different results in Table \ref{tab:accuracy_comparison} (E~+~I) and Table \ref{tab:accuracy_comparison_zevio} (state-of-the-art E~+~I).}

To the best of our knowledge, we are the first to report results on the Event Camera Dataset using all three sensor modalities.
It can be seen that our approach, that uses frames and events, is better in terms of accuracy on almost all the datasets.

\section{Quadrotor flight with an event camera}
\label{sec:quadrotor_flight}

In order to show the potential of our hybrid, frame-and-event--based pipeline in a real scenario, we ran our approach onboard an autonomous quadrotor and used it to fly autonomously in challenging conditions.
We first start by describing in detail the quadrotor platform we built (hardware and software) in section \ref{sec:aerial_platform} before turning to the specific in-flight experiments (section \ref{sec:flight_experiments}).

\subsection{Aerial Platform}
\label{sec:aerial_platform}

\subsubsection{Platform}

\begin{figure}
  \centering

  \subfigure[Quadrotor platform used for the flight experiments.]{\label{fig:quadrotor_platform}\includegraphics[trim={0 0 0 0cm},clip,width=0.515\columnwidth]{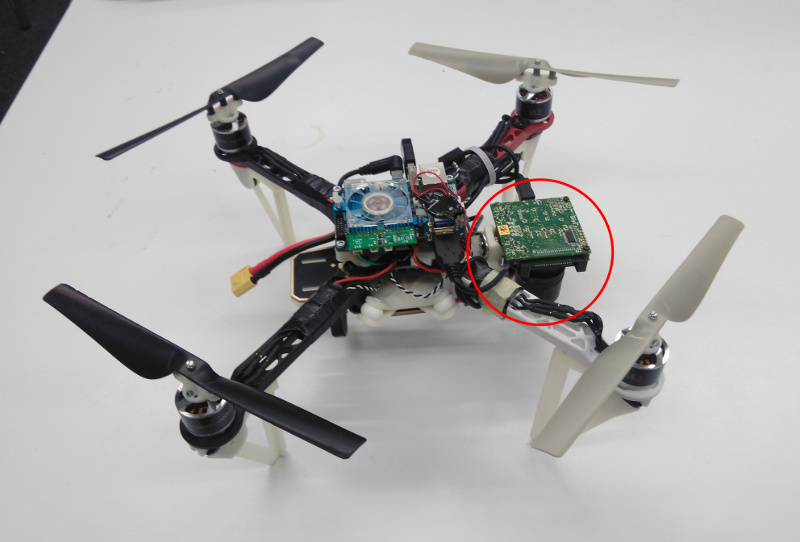}}
  \subfigure[Preview of the room in which we conducted the flight experiments.]{\includegraphics[trim={0 0 0 0cm},clip,width=0.465\columnwidth]{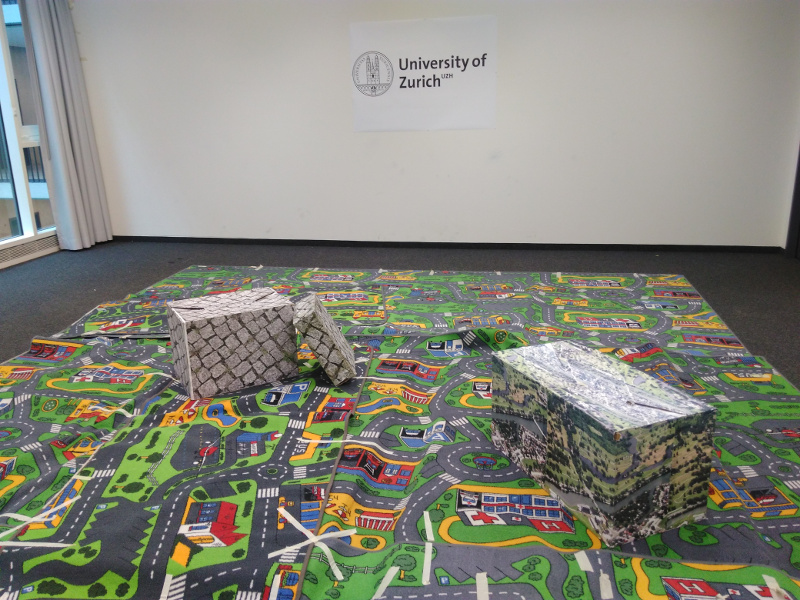}}

  \caption{Quadrotor platform used for the flight experiments, and preview of the flying room.}
\end{figure}

We built our quadrotor from selected off-the-shelf components and custom 3D printed parts (Fig.~\ref{fig:quadrotor_platform}).
Our quadrotor relies on a DJI frame, with RCTimer motors and AR drone propellers.
The electronic parts of our quadrotor comprise a PX4FMU autopilot \cite{Meier12ar}.
In addition, our quadrotor is equipped with an Odroid XU4 computer, which
contains a 2.0 GHz quad-core processor running Ubuntu 14.04 and ROS~\cite{Quigley09icraoss}.
Finally, a DAVIS 240C sensor, equipped with a $70^\circ$ field-of-view lens, is mounted on the front of the quadrotor, looking downwards. 
The sensor is connected to the Odroid computer via an USB 2.0 cable, and transmits events, standard frames, and inertial measurements, which we use to compute the state estimate on the Odroid using our proposed pipeline.
Since the available ROS driver for the DAVIS did not come with an auto-exposure for the standard camera, we implemented an auto-exposure algorithm and made it available open-source for the community to use.\footnote{Available in the DAVIS ROS driver: https://github.com/uzh-rpg/rpg\_dvs\_ros}
It is based on a simple proportional controller that controls the mean image intensity to a desired value (we used a value of $70$ in our experiments).

\vspace{0.45ex}
\subsubsection{Control}

To follow reference trajectories and stabilize the quadrotor, we use the cascaded controllers presented in \cite{Faessler16jfr}.
The high-level controller running on the Odroid includes a position controller and an attitude controller, while the low-level controller on the PX4 contains a body rate controller.
The high-level controller takes a reference trajectory as input and computes desired body rates that are sent to the low-level controller.
The low-level controller, in turn, computes the desired rotor thrusts using a feedback linearizing control scheme with the closed-loop dynamics of a first-order system.
Details of the controllers can be found in \cite{Faessler16jfr}.

\subsection{Flight Experiments}
\label{sec:flight_experiments}

We present \Rebuttal{three} flight experiments that demonstrate that our system is able to fly a quadrotor in challenging conditions: (i) flying indoors while switching on and off the light (which is challenging because of the abrupt large change of illumination caused by the switching of the light and the very low light present in the room after the artificial light is turned off); (ii) while performing fast circles in a
low-lit room; \Rebuttal{(iii) hovering at the same position (which is challenging for the event camera because close to no motion).}
In the first case, when the light is off, the standard frames are completely black.
In the second one, the speed of the quadrotor induces severe motion blur on the standard frames. Nevertheless, in both cases, the events are left unaffected and our pipeline is able to successfully exploit them to provide robust state estimation.
\Rebuttal{
In the third case, instead, there is almost no motion, which makes it difficult for the event camera to track reliable features. Nevertheless, the frames are left unaffected and our pipeline is able to successfully exploit them to provide robust state estimation.
}

These \Rebuttal{three} experiments are best appreciated in video attachment.\footnote{\url{http://rpg.ifi.uzh.ch/ultimateslam.html}}

\begin{figure*}[htb]
  \centering

\begin{tikzpicture}
  \node(img1)  {
    \includegraphics[trim={0 0.0cm 0 0.08cm},clip,width=0.245\textwidth]{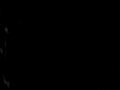}
    \includegraphics[trim={0 0 0 0.08cm},clip,width=0.245\textwidth]{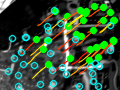}
    \includegraphics[trim={0 0 0 0.08cm},clip,width=0.245\textwidth]{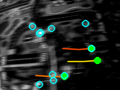}
  };\hspace{-0.1cm}
\node[left=of img1, node distance=0cm, rotate=90, anchor=center,yshift=-0.7cm,font=\color{black}] {\textbf{Standard Frames}};
\end{tikzpicture}\vspace{-0.23cm}
\begin{tikzpicture}
  \node (img1)  {
    \subfigure[Low-light.]{\label{fig:lightoff_circle_events}\includegraphics[trim={0 0cm 0 0.08cm},clip,width=0.245\textwidth]{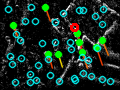}}
    \subfigure[Good lighting, moderate speed.]{\label{fig:normal_condition_circle_events}\includegraphics[trim={0 0 0 0.08cm},clip,width=0.245\textwidth]{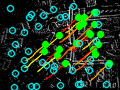}}
    \subfigure[Motion blur.]{\label{fig:motion_blur_circle_events}\includegraphics[trim={0 0 0 0.08cm},clip,width=0.245\textwidth]{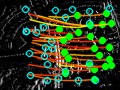}}
    };
\node[left=of img1, node distance=0cm, rotate=90, anchor=center,yshift=-0.6cm,font=\color{black}] {\textbf{Event Frames}};
\end{tikzpicture}
\vspace{-0.4cm}

    \caption{Example feature tracks in various conditions, on the standard frames (top row) and the virtual event frames (bottom row). Every column corresponds to the same timestamp, a frame from the top row has a corresponding event frame on the bottom row. The green solid dots are persistent features, and the blue dots correspond to candidate features. The tracks are shown as colored lines.}

  \label{fig:circle_experiment}
\end{figure*}

\begin{figure}[t]
  \centering     
  \begin{minipage}[b][][s]{\columnwidth}
    \centering
  \subfigure[Top view.]{\includegraphics[trim={0 0 0 0cm},clip,width=0.365\columnwidth]{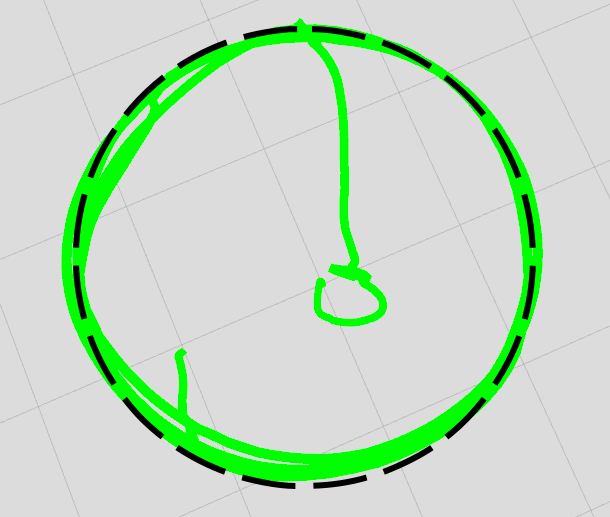}}
  \subfigure[Perspective view.
]{\includegraphics[trim={0 0 0 0cm},clip,width=0.492\columnwidth]{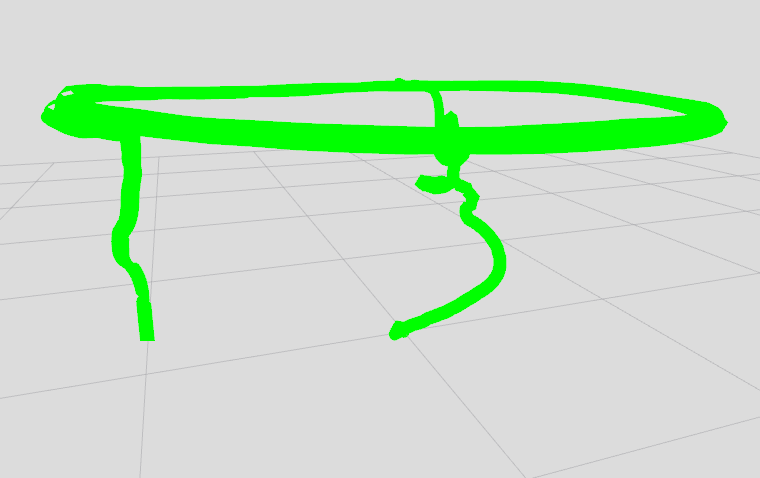}}
  \end{minipage}
  \caption{Experiment 1: switching the light off and on.  The trajectory estimated by our pipeline is the green line. The commanded trajectory is the superimposed black dashed line.}
  \label{fig:top_circle_exp1}
\end{figure}

\begin{figure}[t]
  \centering     
  \begin{minipage}[b][][s]{\columnwidth}
    \centering
  \subfigure[Top view.]{\includegraphics[trim={0 0 0 0.5cm},clip,width=0.36\columnwidth]{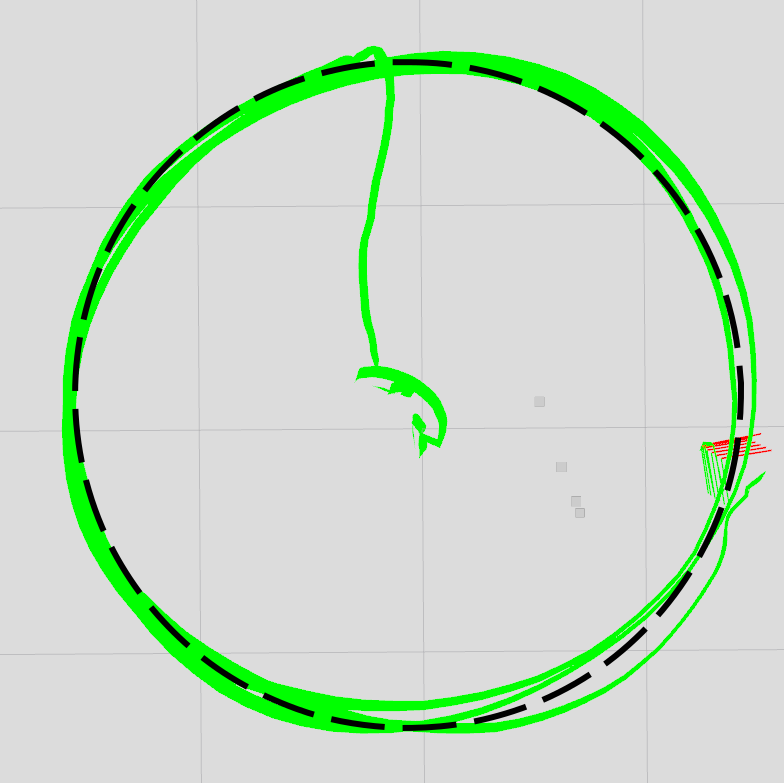}}
  \subfigure[Perspective view.]{\includegraphics[trim={0 0 0 0cm},clip,width=0.48\columnwidth]{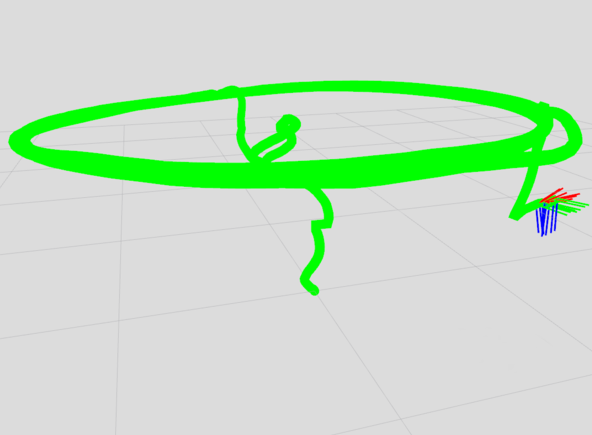}}
  \end{minipage}
    \caption{Experiment 2: Fast circles in a low-lit room. The trajectory estimated by our pipeline is the green line. The commanded trajectory is the superimposed black dashed line. }
  \label{fig:top_circle_exp2}
\end{figure}

\vspace{0.45ex}
\subsubsection{Switching the light off and on, in flight}
In this experiment, we pushed our pipeline to the limit by outright switching the room light off while autonomously flying in circles.
The only remaining light was residual light coming from the windows (very little light, but still enough for the event camera to work).
The standard frames become completely black when the light goes off (top frame in Fig.~\ref{fig:lightoff_circle_events}), making them useless for state estimation.
By contrast, the events still carry enough information (albeit noisier) to allow  reasonable feature tracks (bottom frame Fig.~\ref{fig:lightoff_circle_events}).
Switching the light off effectively forces the pipeline to rely only on events and inertial measurements.
Note that the abrupt illumination change caused by switching the lights on and off makes almost every pixel fire events.
Although we do not explicitly handle this particular case, in practice we observed no substantial decrease in accuracy when this occurs as features are quickly re-initialized.

The trajectory flown by the quadrotor is shown in Fig.~\ref{fig:top_circle_exp1}.

\vspace{0.45ex}
\subsubsection{Fast Circles in a Low-lit Room}

In this experiment, the quadrotor autonomously flies a circular trajectory with increasing speed
in a closed room with little light (Fig.~\ref{fig:eye_catcher}); we carried this experiment during the night and set a low lighting in the room.
The circular trajectory commanded to the quadrotor is parametrized by its radius and the desired angular velocity.
We set the angular velocity to $1.4$~rad/s on a circle of $1.2$~m radius, corresponding to a top linear velocity of $1.68$~m/s.
The circle height was $1.0$~m.
At this speed and height, the optical flow generated on the image plane amounts to approximately $340$~pixels/s.

While the speed remains moderate at the beginning of the trajectory (below $1.2$~m/s), standard frames do not suffer from motion blur and our pipeline indeed tracks features in both the standard frames and the event frames (cf. top and bottom frames in Fig.~\ref{fig:normal_condition_circle_events}, respectively).
Nevertheless, as soon as the speed increases, the standard frames start to suffer from severe motion blur, as shown in the top frame of Fig.~\ref{fig:motion_blur_circle_events}, and the number of features tracked in the standard frames significantly decreases.
Conversely, the events allow synthesizing motion-free virtual event frames, which, in turn, allow  keeping reliable feature tracks (bottom frame in Fig.~\ref{fig:motion_blur_circle_events}).

In Fig.~\ref{fig:top_circle_exp2}, both the desired and estimated trajectories are shown for comparison.
Interestingly, the right side of the trajectory is slightly noisier than the left side.
This turns out to match well with the light configuration in the room: the left side of the room was indeed more illuminated than the right side (visible in Fig.~\ref{fig:eye_catcher}).
This is coherent with the quantitative experiments presented in section \ref{sec:quantitative_evaluation}: the increase of the quality of the standard frames on the room side with more light correlates directly to an increase of accuracy of the pipeline.

\vspace{0.45ex}
\subsubsection{Hovering}

\Rebuttal{We also provide qualitative experiments to show how our pipeline performs  close to no-motion conditions, typically encountered when a drone is hovering.}

\Rebuttal{
First, we command the drone to hover while using the events-only pipeline and observe that the state estimate drifts.
We then command the drone to hover while using both the images and the event frames, and observe that the drone successfully keeps its position with no noticeable drift.}

\Rebuttal{
The difference lies in that features are tracked successfully on the standard frames, 
while they are lost on the event frames. This is because, unlike standard cameras, the 
appearance of the features tracked by the event camera may change “drastically” with 
the direction of the motion, which is exactly what happens when hovering: vibrations 
induce frequent changes of motion direction, which reduce the length of the feature 
tracks from the event streams, leading to increased drift. 
}

\section{Conclusions}
\label{sec:conclusions}
We introduced the first hybrid pipeline that fuses events, standard frames, and inertial measurements to yield robust and accurate state estimation.
We also reported results using these three sensing modalities on the Event Camera Dataset \cite{Mueggler17ijrr} and demonstrated an accuracy boost of 130~\% compared to using only events plus IMU, and a boost of 85~\% compared to using only standard frames plus IMU.
Furthermore, we successfully integrated the proposed pipeline for state estimation onboard a computationally-constrained quadrotor and used it to realize, to the best of our knowledge, the first closed-loop flight of a quadrotor using an event camera.
Finally, in a set of specific experiments, we showed that our hybrid pipeline is able to leverage the properties of the standard camera and the event camera to provide robust tracking when flying in multiple conditions, such as hovering, flying in fast circles or flying in a low-lit room.

\addtolength{\textheight}{-11cm}   


\bibliographystyle{IEEEtran} 
\bibliography{all}

\addtolength{\textheight}{0cm}   

\begin{figure*}[t]
\section*{Appendix}
\centering     

\subfigure[\texttt{boxes\_6dof}]{\includegraphics[trim={0 0 0 0cm},clip,width=0.48\textwidth]{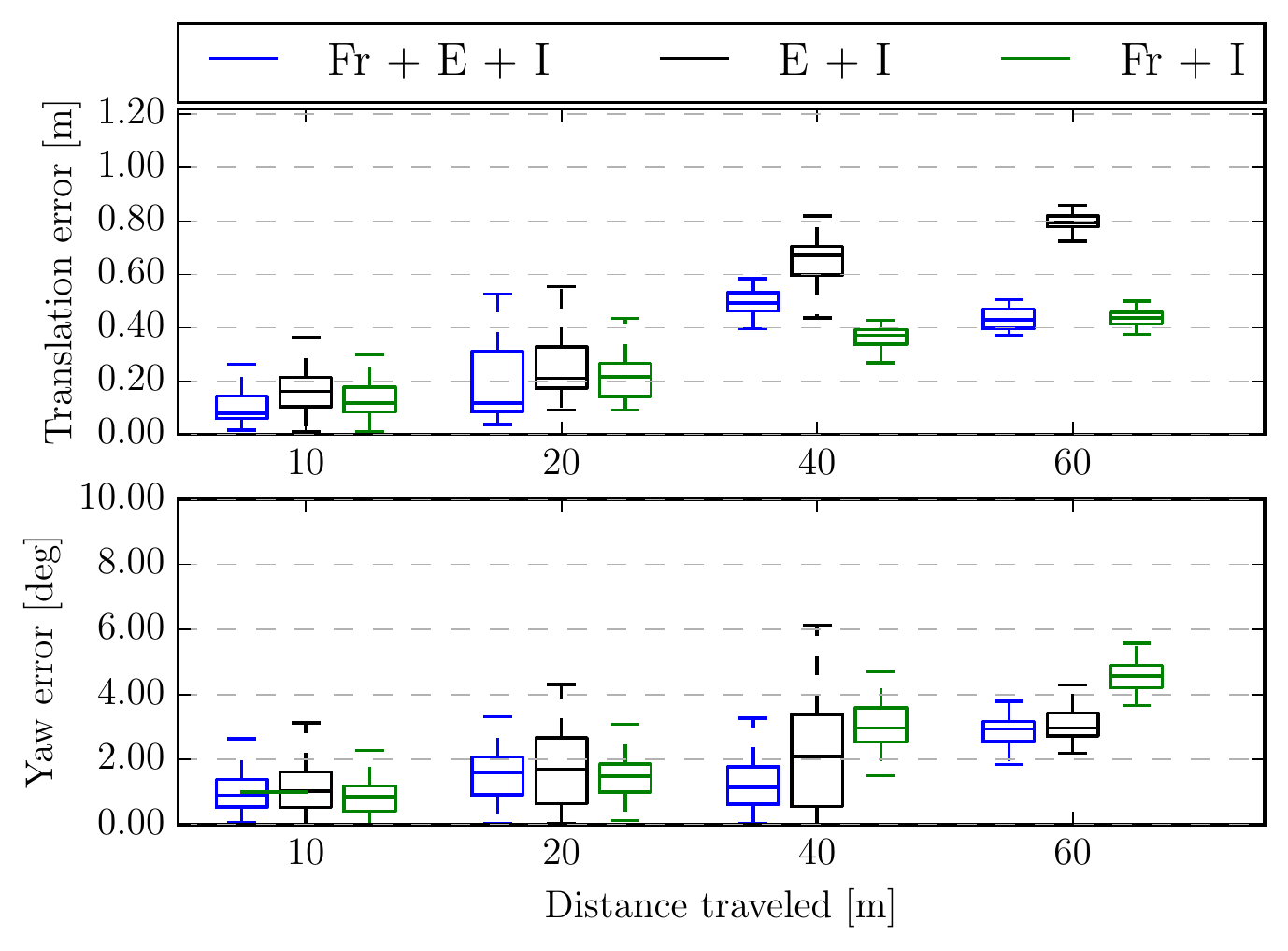}}
\subfigure[\texttt{boxes\_translation}]{\includegraphics[trim={0 0 0 0cm},clip,width=0.48\textwidth]{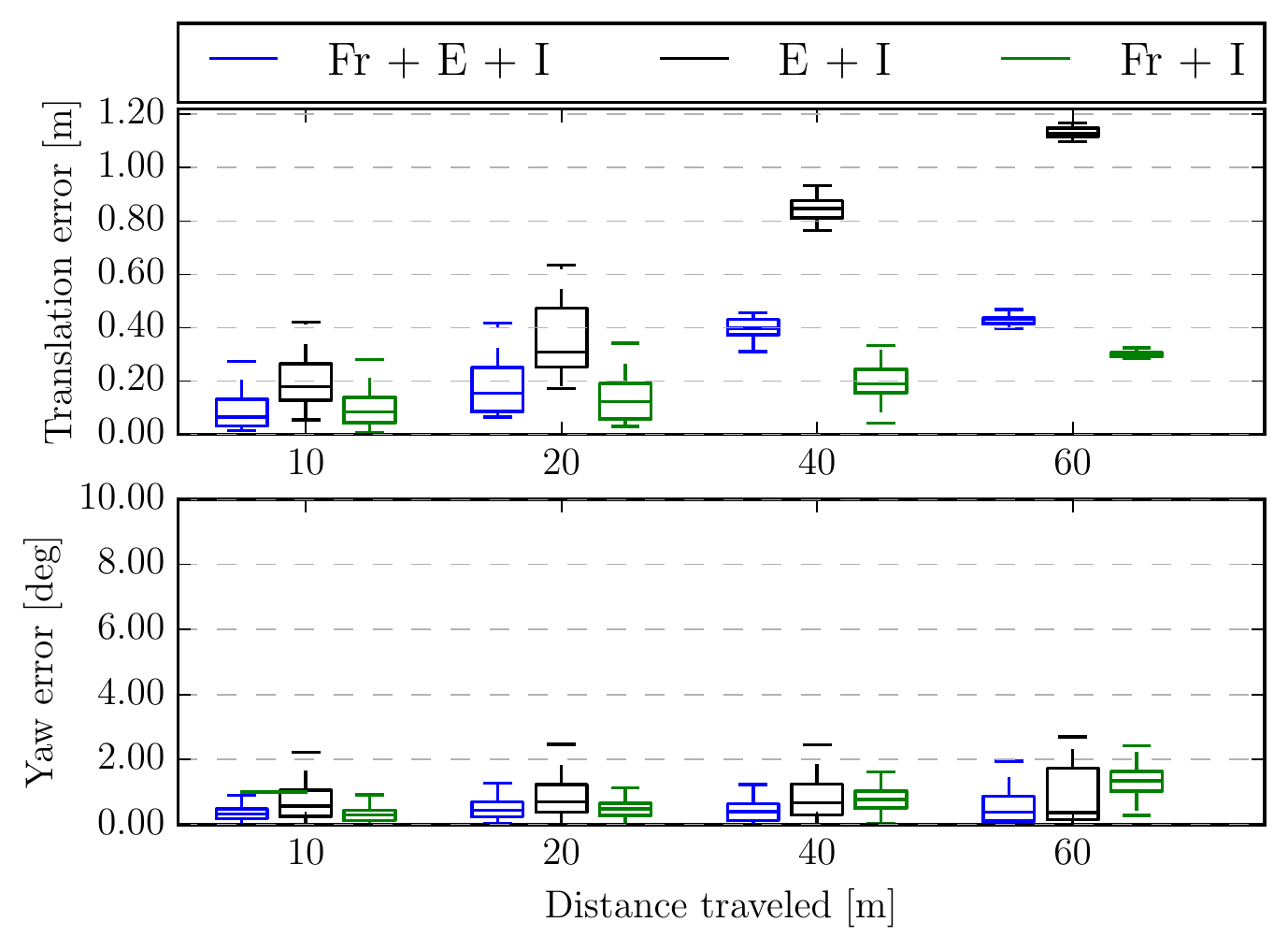}}

\subfigure[\texttt{dynamic\_6dof}]{\includegraphics[trim={0 0 0 0cm},clip,width=0.48\textwidth]{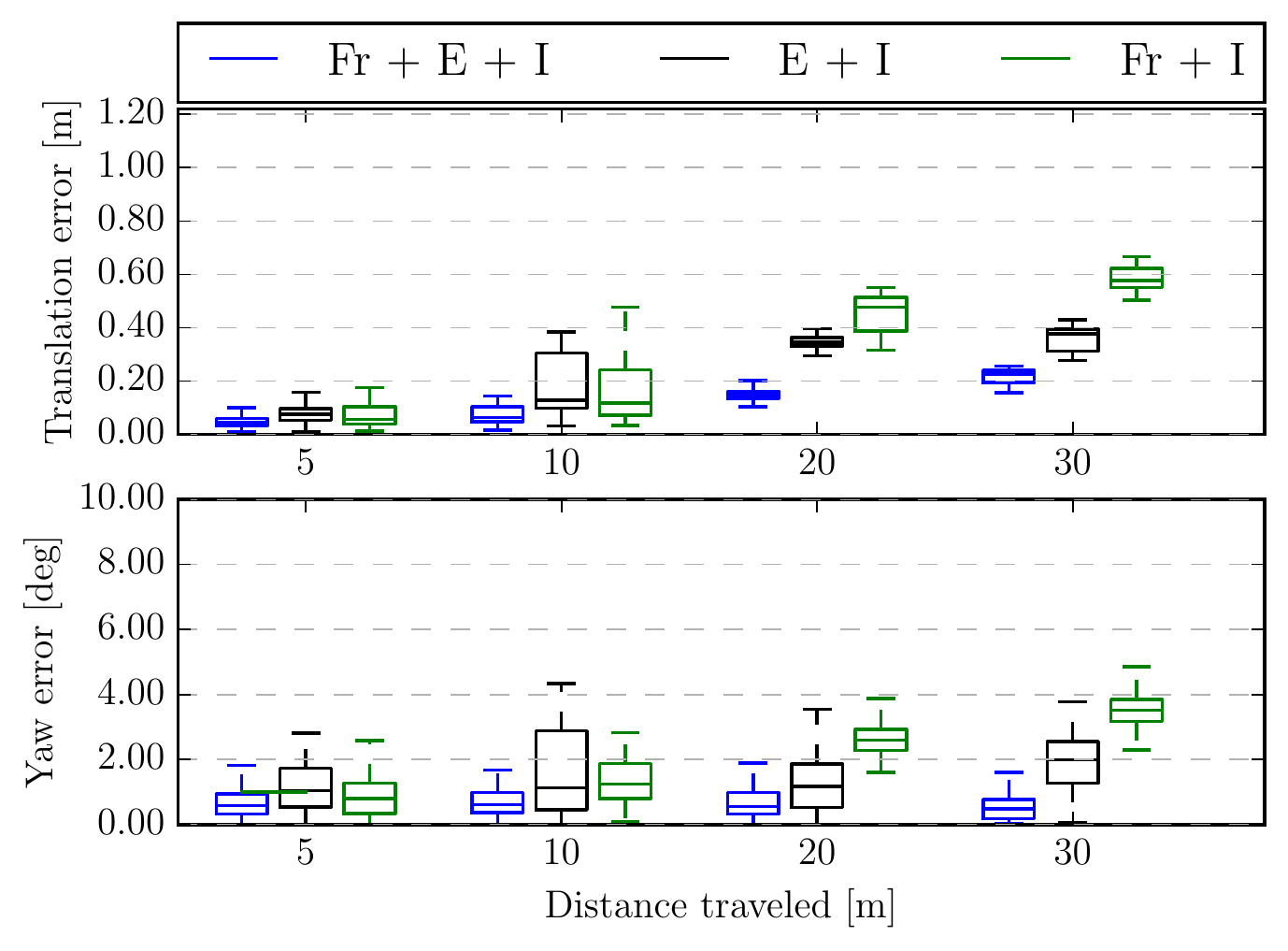}}
\subfigure[\texttt{dynamic\_translation}]{\includegraphics[trim={0 0 0 0cm},clip,width=0.48\textwidth]{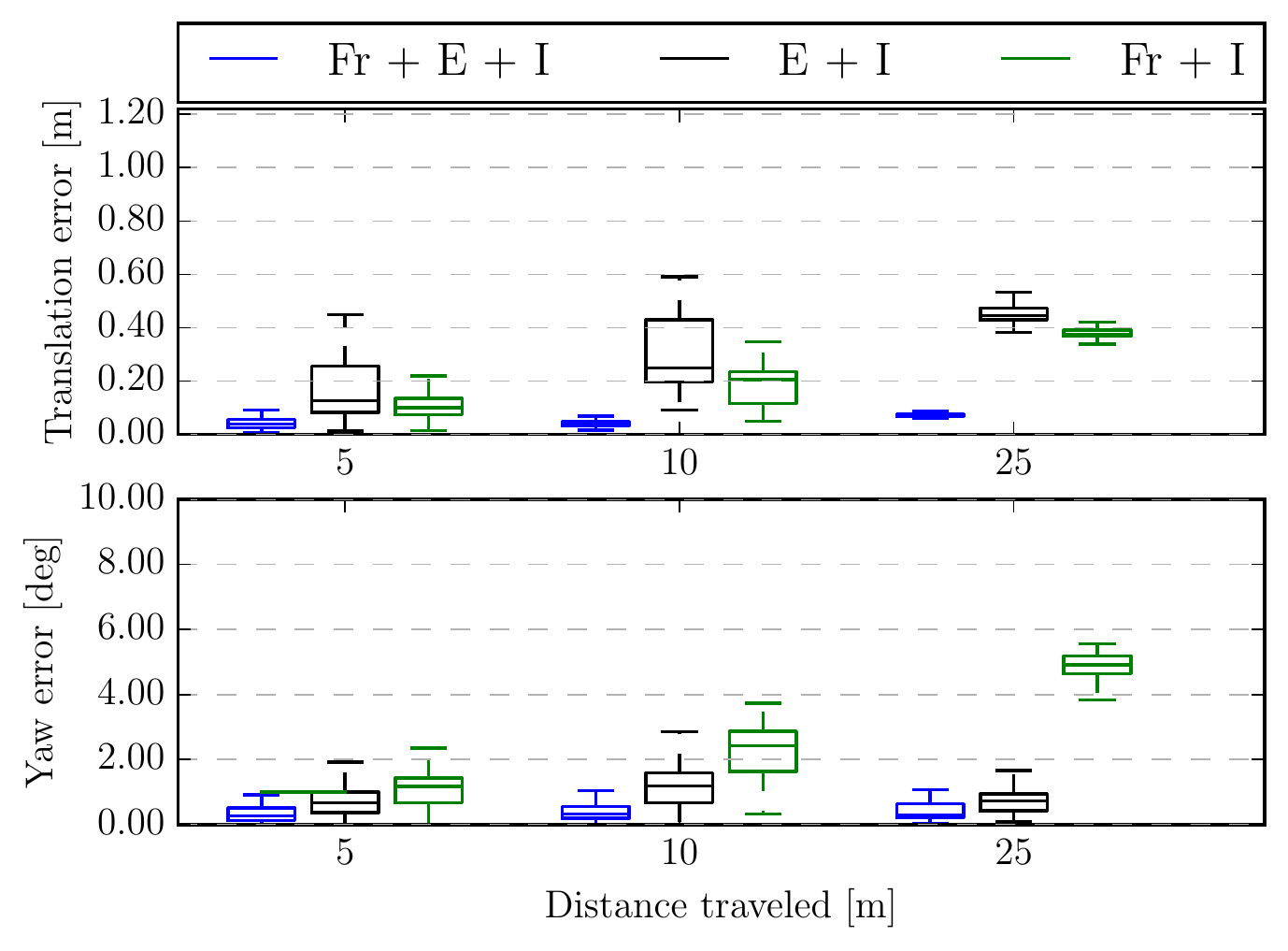}}

\subfigure[\texttt{hdr\_boxes}]{\includegraphics[trim={0 0 0 0cm},clip,width=0.48\textwidth]{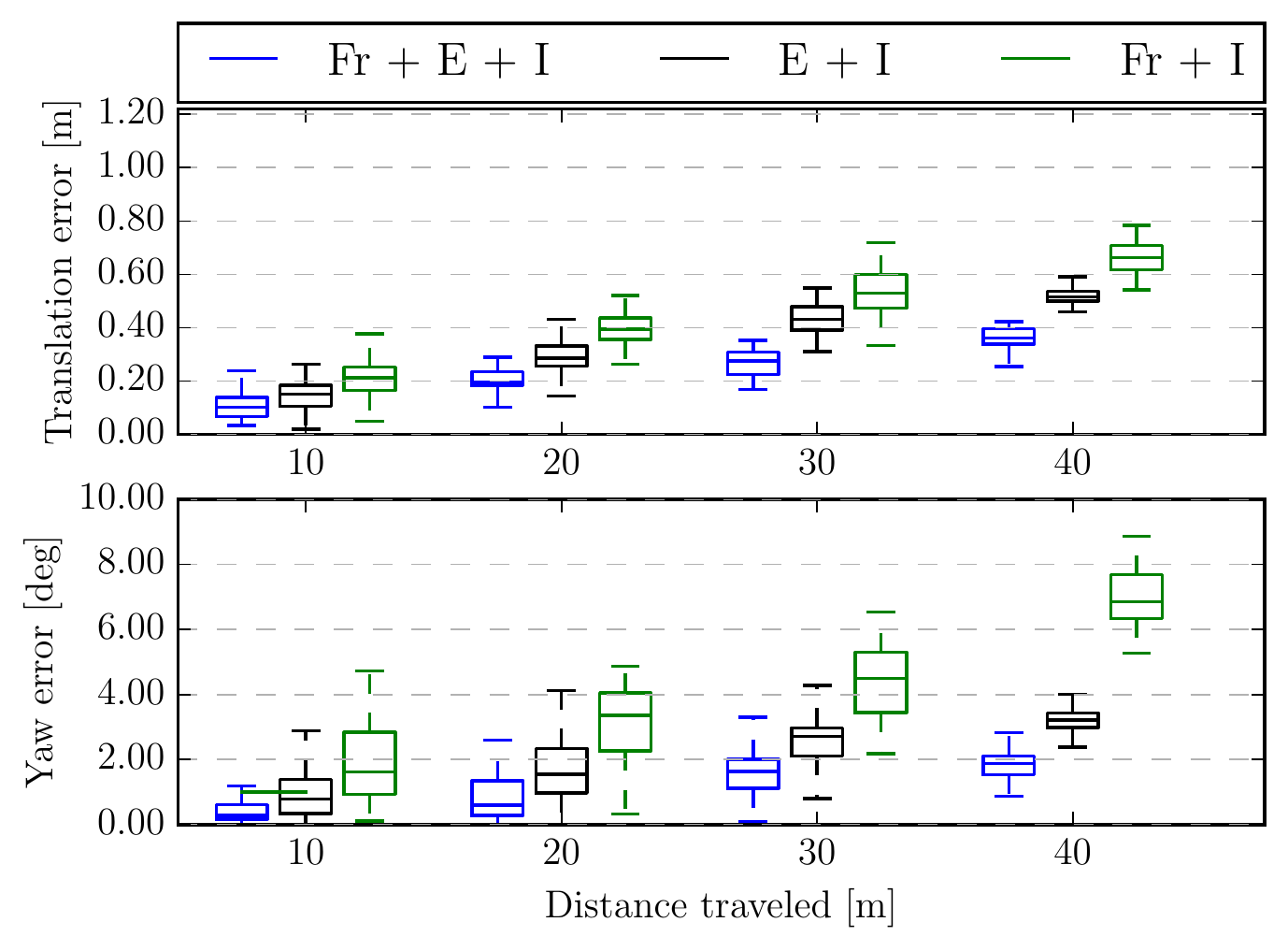}}
\subfigure[\texttt{hdr\_poster}]{\includegraphics[trim={0 0 0 0cm},clip,width=0.48\textwidth]{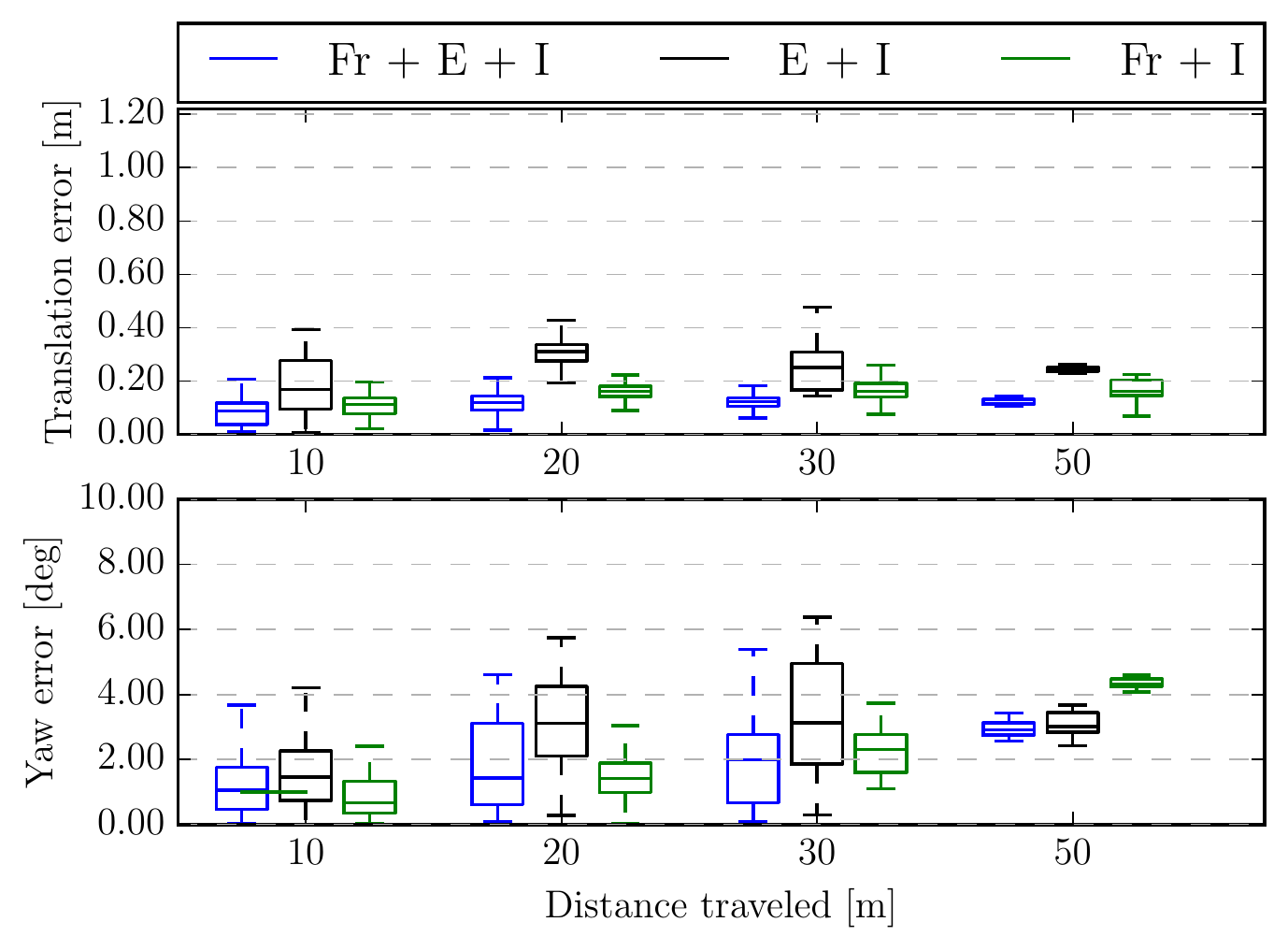}}

\caption{Detailed comparison of the pipeline performance on the Event Camera Dataset, while using standard frames (Fr), events (E), and IMU (I); events and IMU; and images and IMU.}

\label{fig:results_full}
\end{figure*}

\begin{figure*}[t]
\centering     

\subfigure[\texttt{poster\_6dof}]{\includegraphics[trim={0 0 0 0cm},clip,width=0.48\textwidth]{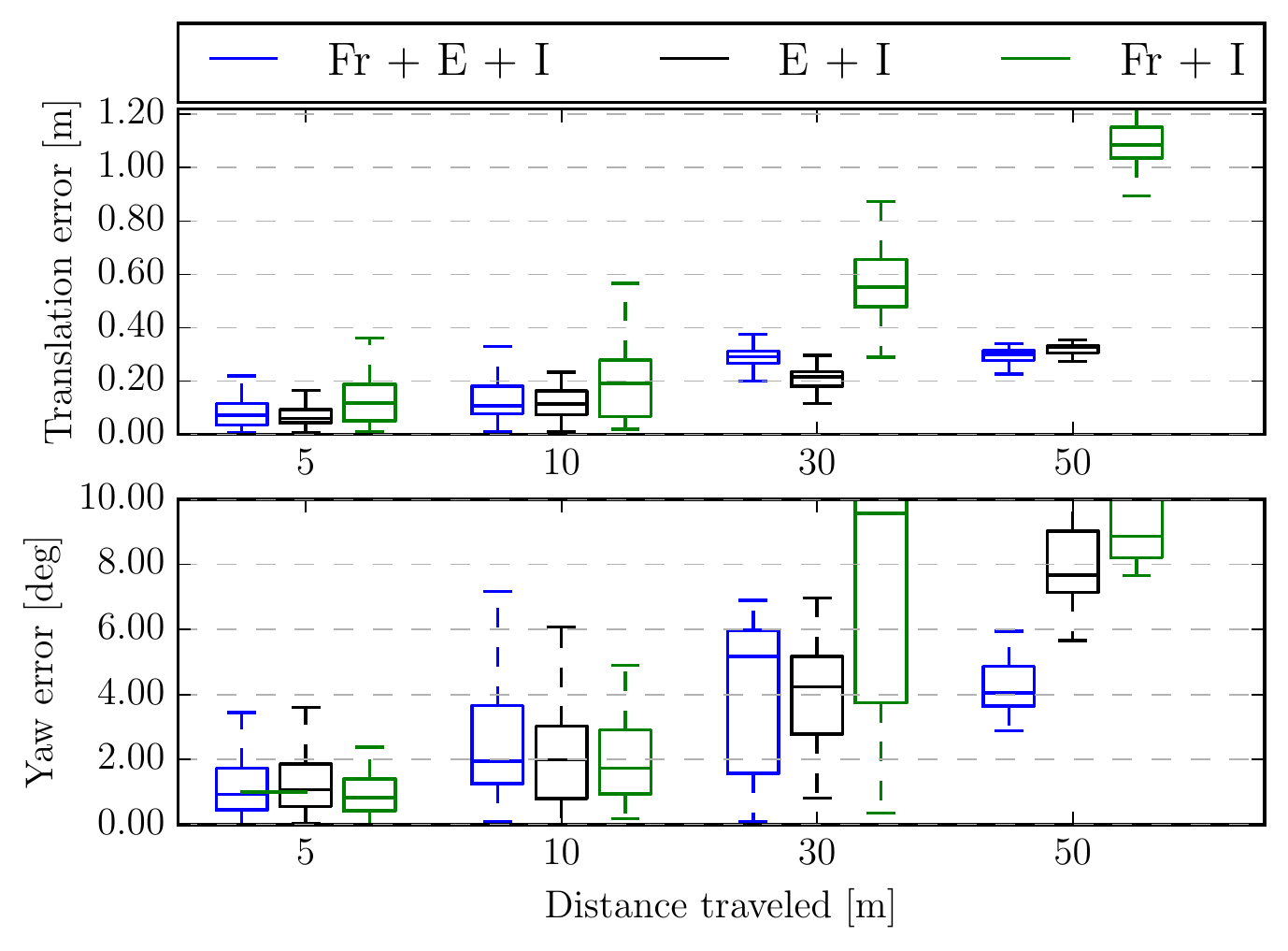}}
\subfigure[\texttt{poster\_translation}]{\includegraphics[trim={0 0 0 0cm},clip,width=0.48\textwidth]{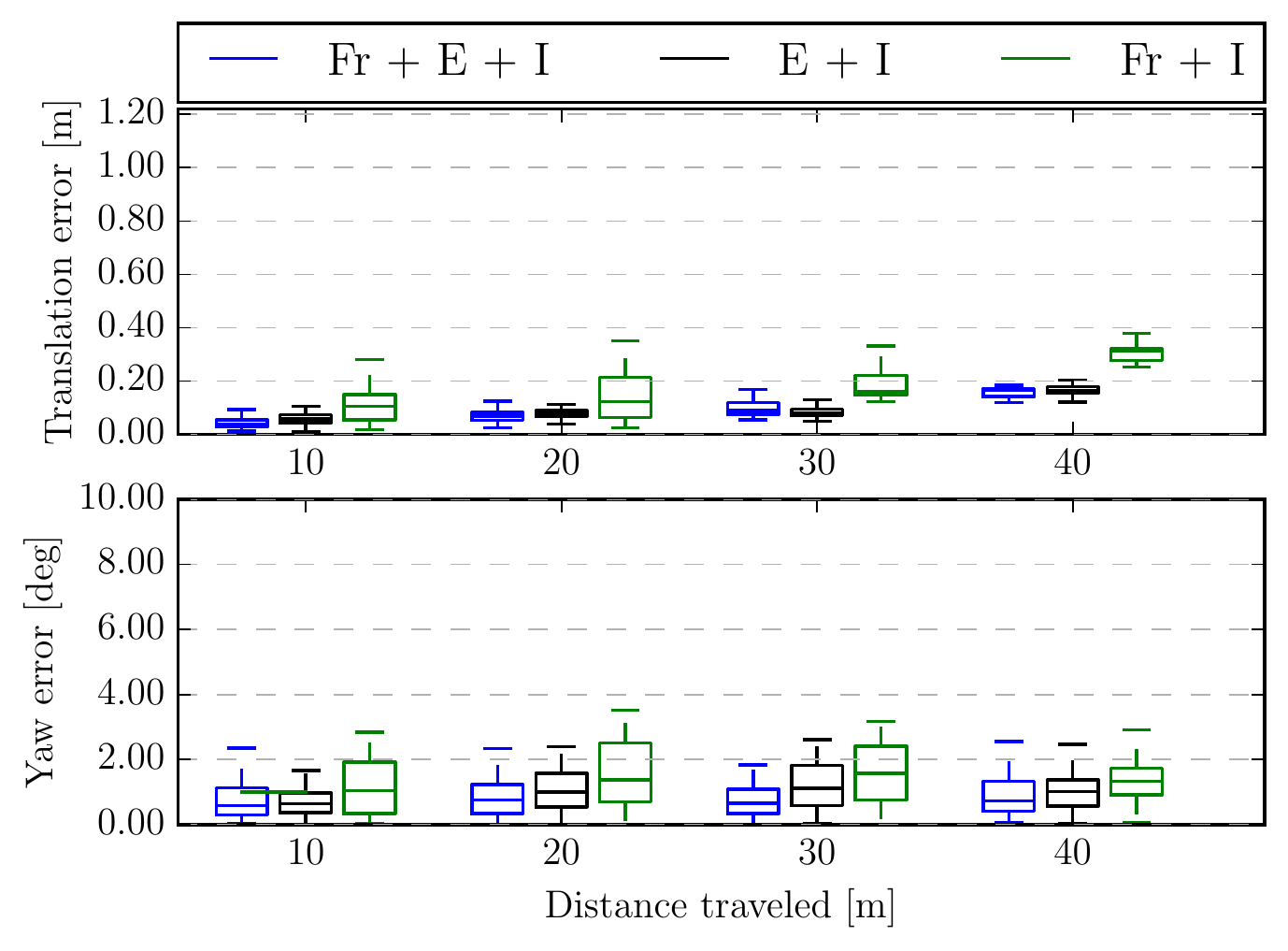}}

\subfigure[\texttt{shapes\_6dof}]{\includegraphics[trim={0 0 0 0cm},clip,width=0.48\textwidth]{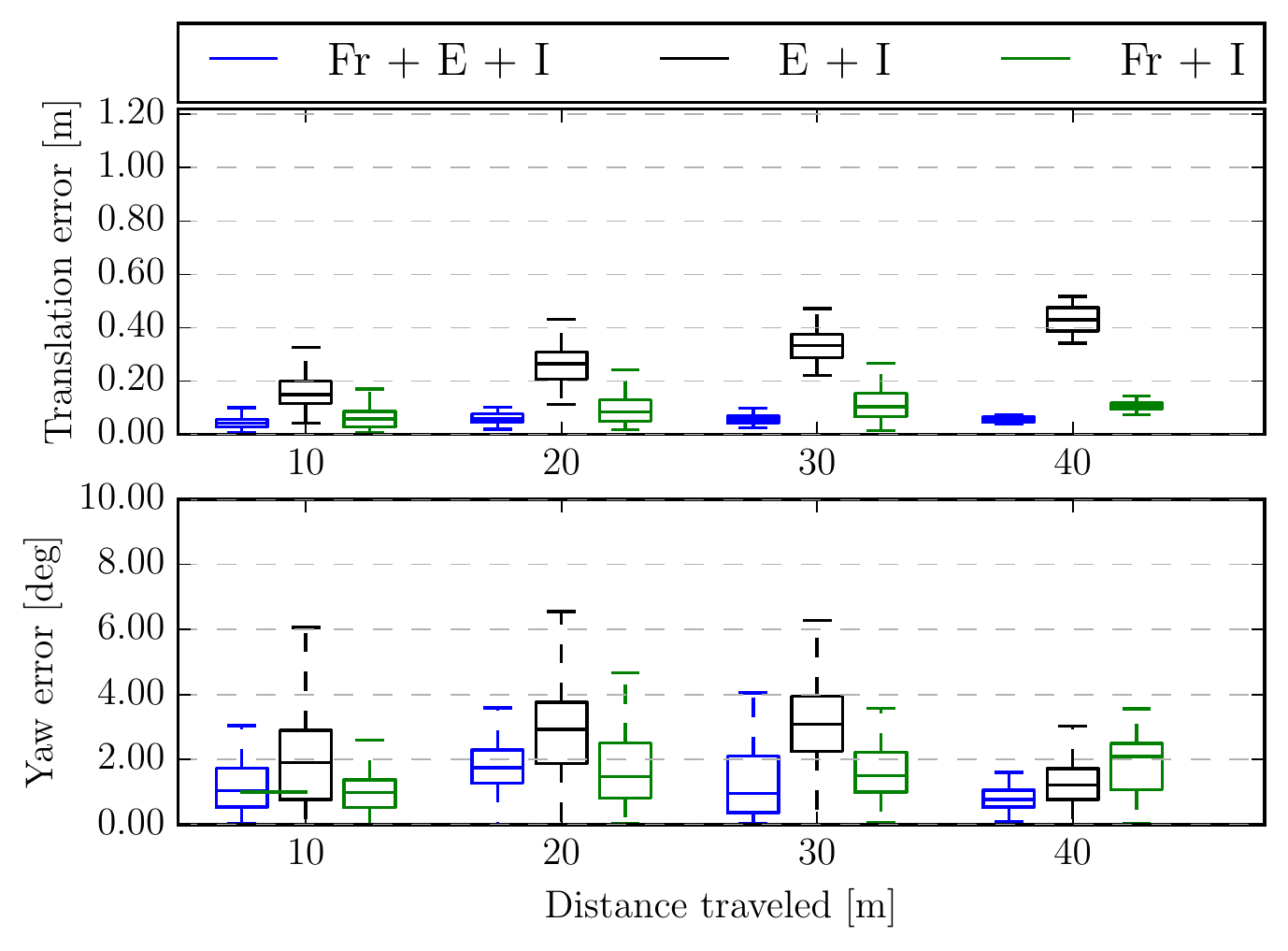}}
\subfigure[\texttt{shapes\_translation}]{\includegraphics[trim={0 0 0 0cm},clip,width=0.48\textwidth]{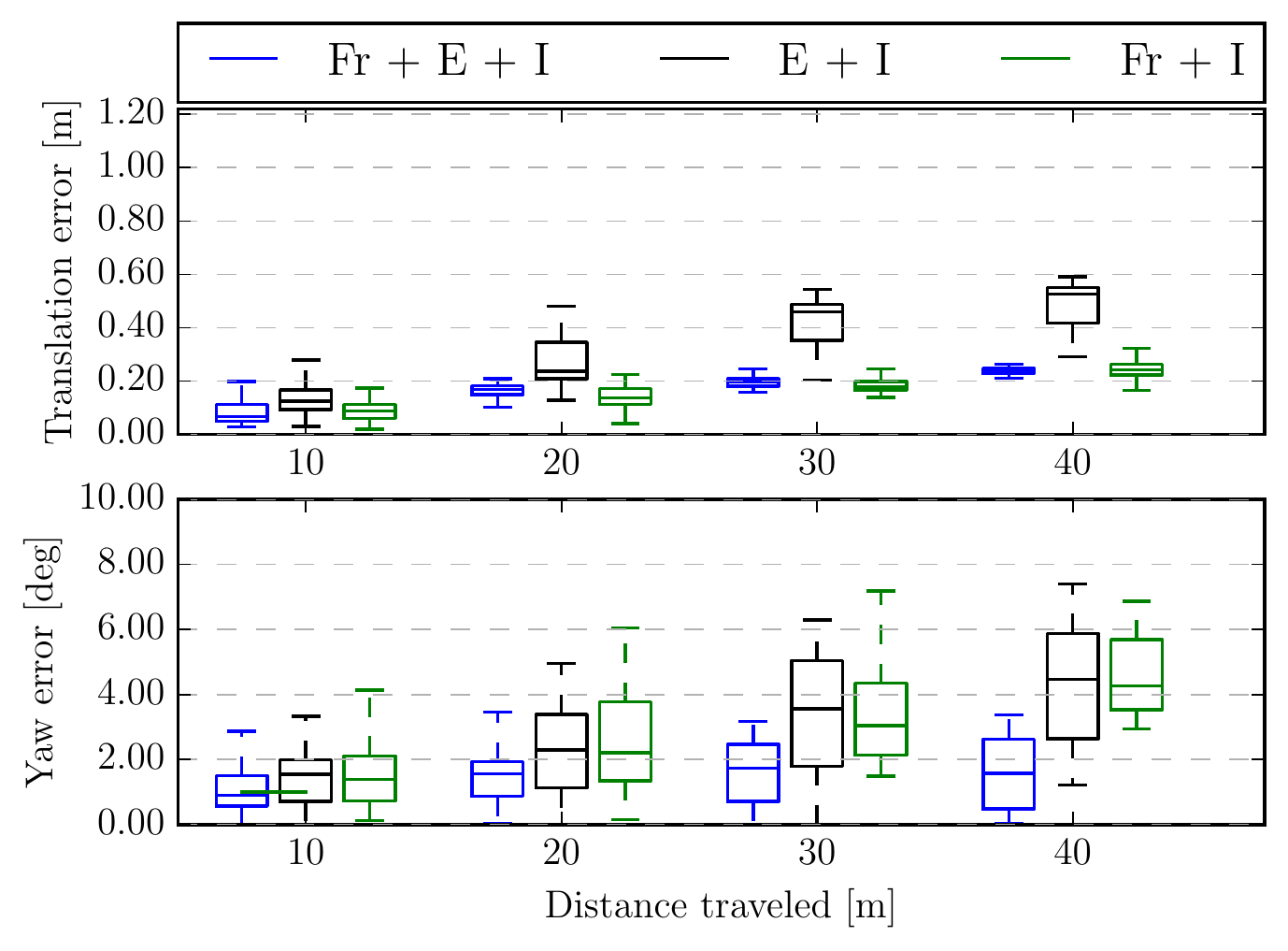}}

\caption{(Continuation) Detailed comparison of the pipeline performance on the Event Camera Dataset, while using standard frames (Fr), events (E), and IMU (I); events and IMU; and images and IMU.}

\label{fig:results_full}
\end{figure*}

\end{document}